\documentclass[runningheads]{llncs}
\usepackage{graphicx}
\usepackage{amsfonts, amsmath}

\begin{document}
\title{Unsupervised anomaly localization using VAE and beta-VAE}
%
%
\author{Leixin Zhou\inst{1} \and
Wenxiang Deng\inst{1} \and
Xiaodong Wu\inst{1}}
%
\institute{Department of ECE, University of Iowa, Iowa city IA 52242, USA
\email{leixin-zhou@uiowa.edu}\\}
%
\maketitle              
\begin{abstract}
Variational Auto-Encoders (VAEs) have shown great potential in the unsupervised learning of data distributions. An VAE trained on normal images is expected to only be able to reconstruct normal images, allowing the localization of anomalous pixels in an image via manipulating information within the VAE ELBO loss. The ELBO consists of KL divergence loss (image-wise) and reconstruction loss (pixel-wise). It is natural and straightforward to use the later as the predictor. However, usually local anomaly added to a normal image can deteriorate the whole reconstructed image, causing segmentation using only naive pixel errors not accurate.  Energy based projection was proposed to increase the reconstruction accuracy of normal regions/pixels, which achieved the state-of-the-art localization accuracy on simple natural images. Another possible predictors are ELBO and its components  gradients with respect to  each pixels. Previous work claimed that KL gradient is a robust predictor. In this paper, we argue that the energy based projection in medical imaging is not as useful as on natural images. Moreover, we observe that the robustness of KL gradient  predictor totally depends on the setting of the VAE and dataset. We also explored the effect of the weight of KL loss within beta-VAE and predictor ensemble in anomaly localization.
\end{abstract}
\section{Introduction}
Automating anomaly detection in medical imaging with artificial intelligence has gained popularity and interest in recent years. Indeed, the analysis of images to localize potential abnormality seems well suited to supervised computer vision algorithms. However these solutions remain data hungry and require knowledge transfer from human to machine via image annotations. Furthermore, the classification in a limited number of user-predefined categories such as healthy, tumor and so on, will not generalize well if a previously unseen anomaly appears. For visual inspection, a better-suited task is unsupervised anomaly detection, in which the localization of the abnormality must be done only via prior knowledge of normal samples.
\par
From a statistical point of view, an anomaly may be seen as \textit{ an observation that deviates so much from other observations as to arouse suspicion that it was generated by a different mechanism}\cite{hawkins1980identification}. In this setting, deep generative models such as Variational AutoEncoders (VAEs)\cite{kingma2013auto} and $\beta$-VAE \cite{higgins2017beta}, are especially interesting because they are capable to infer possible sampling mechanisms for a given dataset. The VAE jointly
learns an encoder model, which compresses input samples into a low dimensional space, and a decoder, which decompresses the low dimensional samples into the original input space, by simultaneously minimizing the distance between the input of the encoder and the output of the decoder, and the distance between latent distribution and a prior distribution (usually Gaussian). The output decompressed sample for a given input is often called the \textit {reconstruction}, and is used as some sort of projection of the input on the support of the normal data distribution, usually called the \textit{normal manifold}. In most unsupervised anomaly localization methods based on VAE,
models are trained on normal data and anomaly localization is then performed using a distance metric between the input sample and its reconstruction \cite{bergmann2019mvtec,baur2018deep,chen2018unsupervised}. The localization part in those studies is solely based on the reconstruction error, thus outlining regions as suspicious if they cannot be adequately reconstructed by the model. One obvious deficiency is that the capability of a VAE to reconstruct anomalies is by design tightly coupled to the expressiveness (size and configuration) of the latent space. Then to further improve the localization performance, there are at least two branches to explore: 1) searching for other predictors that are not highly dependent on the modeling capacity of VAE; 2)  principally improving the projection quality such that normal regions in the projected normal image are the same as that in the input image. The loss of VAE, called the evidence lower bound (ELBO), consists of two parts: reconstruction loss and Kullback-Leibler (KL)-divergence loss. Zimmerer \textit{et al}. \cite{zimmerer2019unsupervised,zimmerer2019case} found that KL loss gradient with respect to input is one robust predictor. For the latter branch, instead of using the VAE reconstruction, Dehaene \textit{et al}. \cite{dehaene2020iterative} proposed to iteratively project the abnormal data to the normal manifold more accurately by optimizing a specific energy function. For natural images, they found that their reconstruction error based method outperforms \cite{zimmerer2019unsupervised,zimmerer2019case} significantly. 
\par
In this paper, we argue that the performance of different predictors are highly dependent on the VAE settings, e.g. the size of latent space and the weight of KL loss (VAE becomes $\beta$-VAE). We also test the energy minimization projection based method \cite{dehaene2020iterative} in the medical imaging (T2 MRI Brain images) scenario, and found that it is not as powerful as on simple natural images.
\section{Method}
\subsection{VAE and $\beta$-VAE}
In unsupervised anomaly detection, the only available data during training are samples $\mathbf{x}$ from a normal dataset $\mathbb{X} \subset \mathbb{R}^d$. In a generative setting, we assume the existence of a probability function of density $q$, having its support on all $\mathbb{R}^d$. The generative objective is to model an estimate of $q$, from which one can obtain new samples close to the dataset. 
\par
Popular deep generative models are generative adversarial networks (GAN) \cite{goodfellow2014generative} and VAE. The advantages of GANs are that they can generate sharp and realistic samples, as a discriminator is trained simultaneously to guide the generator. However, disadvantages of GANs are that they are notoriously difficult to train \cite{goodfellow2016nips}, and suffer from mode collapse, meaning that they have the tendency to only generate a subset of the original dataset. This can be problematic for anomaly detection, in which we do not want some subset of the normal data to be considered as anomalous \cite{bergmann2019mvtec}. Recent works such as \cite{thanh2019improving} propose substantial upgrades, however other works such as \cite{ravuri2019classification} still supports that GANs have more trouble than other generative models to cover the whole distribution support.
\par 
Another deep generative model is VAE, which consists of an encoder and a decoder. The decoder, similar to a GAN generator, tries to approximate the conditional dataset distribution $p(\mathbf{x|z})$ on a simple latent variables prior $p(\mathbf{z}),\mathbf{z}\in \mathbb{R}^l$. We would like to maximize the estimate $p(\mathbf{x})=\int p(\mathbf{x|z}) p(\mathbf{z}) dz$ on the dataset. To make the learning tractable, importance sampling by introducing density functions $q(\mathbf{z|x})$ output of an encoder is utilized, and the variational evidence lower bound (ELBO) $\mathcal{L}$ can be deduced as:
\begin{equation}
	\begin{split}
		\log p(\mathbf{x})&=\log \mathbb{E}_{\mathbf{z}\sim q(\mathbf{z|x})} \frac{p(\mathbf{x|z})p(\mathbf{z})}{q(\mathbf{z|x})} \\
		&\geq  \mathbb{E}_{\mathbf{z}\sim q(\mathbf{z|x})} \log p(\mathbf{x|z}) - D_{KL}(q(\mathbf{z|x}) || p(\mathbf{z})) = ELBO = -\mathcal{L}
	\end{split}
\end{equation}
$\mathcal{L}$, the opposite of ELBO, is utilized as the loss function of VAE for training. VAEs are known to produce blurry reconstructions and generations. The advantages are that VAEs probably do not suffer the mode collapse problem \cite{razavi2019generating} and VAEs can generate projection of new input to the training dataset manifold in one forward pass, without need of iterative optimization if using GANs \cite{schlegl2017unsupervised}. $\beta$-VAEs share all the merits with VAEs, and its loss function is formulated as:
\begin{equation}
	- \mathbb{E}_{\mathbf{z}\sim q(\mathbf{z|x})} \log p(\mathbf{x|z}) + \beta \cdot D_{KL}(q(\mathbf{z|x}) || p(\mathbf{z}))
\end{equation}
By putting more weight ($\beta > 1$) on the KL term, the trained $\beta$-VAEs encourage disentangled factor learning in the latent space.

\subsection{Predictors for Pixel-wise Anomaly Localization}
We will consider that an anomaly is a sample with low probability under our estimation of the normal dataset distribution. The VAE loss, being a lower bound on the density, is a proxy to classify samples between the normal and abnormal categories. To this effect, a threshold $T$ can be defined on the loss function, where anomalous samples with $\mathcal{L}(\mathbf{x}) \geq T$ and normal samples with $\mathcal{L}(\mathbf{x}) < T$. However, according to Nalisnick \textit{et al}. \cite{nalisnick2018deep}, the likelihood of a data point $p(\mathbf{x})$ in deep generative models is not a reliable measure for detecting abnormal samples. Also according to Matsubara \textit{et al}. \cite{matsubara2018anomaly}, the regularization term $\mathcal{L}_{KL}(\mathbf{x})=D_{KL}(q(\mathbf{z|x}) || p(\mathbf{z}))$ has a negative influence in the computation of anomaly scores. They proposed instead an unregularized score $\mathcal{L}_{r}=-\mathbb{E}_{\mathbf{z}\sim q(\mathbf{z|x})} \log p(\mathbf{x|z})$, which is equivalent to the reconstruction loss of a standard autoencoder. Going from anomaly detection to anomaly localization, this reconstruction term becomes crucial to most of existing solutions. Indeed, the inability of the model to reconstruct a given part of an image is used as a way to segment the anomaly, using a pixel-wise threshold on the reconstruction error \cite{bergmann2019mvtec,baur2018deep,chen2018unsupervised}. We call  it a  reconstruction-loss based predictor. However, according to \cite{zimmerer2019case,zimmerer2019unsupervised}, the magnitude of the loss gradient with respect to $\mathbf{x}$, such as $|\frac{\partial \mathcal{L}}{\partial x_i}|$, $|\frac{\partial \mathcal{L}_{KL}}{\partial x_i}|$ and so on, is useful and maybe more robust predictor.  
To clarify the notations, we list out all the predictors and their formulations, as follows.
\begin{itemize}
\item \textbf{``Rec-Error"}: $\mathbb{E}_{\mathbf{z}\sim q(\mathbf{z|x})} \log p(\mathbf{x|z})$. We parameterize $q(\mathbf{z|x})$ as diagonal Gaussian $\mathcal{N}(\mathbf{z}; f_{\mu}(\mathbf{x}), f_{\sigma}(\mathbf{x})^2)$, and parameterize $p(\mathbf{x|z})$ as $\mathcal{N}(\mathbf{x}; g_{\mu}(\mathbf{z}), \mathcal{I})$. During inference, we approximate it as $\log p(\mathbf{x}|f_{\mu}(\mathbf{x}))$, which is basically the pixel-wise L2 distance between input and reconstruction.
\item \textbf{``ELBO-grad"}: $|\frac{\partial \mathcal{L}}{\partial x_i}|$
\item \textbf{``KL-grad"}: $|\frac{\partial \mathcal{L}_{KL}}{\partial x_i}|$
\item \textbf{``Rec-grad"}: $\frac{\partial \mathbb{E}_{\mathbf{z}\sim q(\mathbf{z|x})} \log p(\mathbf{x|z})}{\partial x_i}$
\item \textbf{``Combi"}: $|\frac{\partial \mathcal{L}_{KL}}{\partial x_i}| \odot \mathbb{E}_{\mathbf{z}\sim q(\mathbf{z|x})} \log p(\mathbf{x|z})$
\end{itemize}
\subsection{Improve Performance of different predictors}
For the two classes of predictors, different strategies can be utilized to improve their respective performance on anomaly localization. For ``Rec-Error" predictor, we apply one iterative projection method, similar to adversarial sample generation, to medical imaging and test its effectiveness. For other predictors, we propose to utilize $\beta$-VAE to  capture other balance between latent space information and reconstruction accuracy for better anomaly localization.
\par \noindent
\textbf{Iterative projection for more accurate reconstruction error}
For ``Rec-Error" predictor, the assumption is that the trained VAE has the capability to alter anomalous pixels and keep normal pixels untouched during reconstruction. In other words, for this predictor, the ideal generative model has the following functional:
\begin{eqnarray}\label{eqn:ideal}
	[\text{VAE}_{ideal}(\mathbf{x})]_i = x_i ~~~~~~~~~~&\text{if pixel}~~ i~~ is ~~\text{abnormal} \nonumber\\
	|[\text{VAE}_{ideal}(\mathbf{x})]_i - x_i| \geq \epsilon ~~~~&\text{otherwise}
\end{eqnarray}
where $\epsilon$ is some positive number. However, practical VAEs can not be  guaranteed to have the aforementioned property held, which makes the  ``Rec-Error" predictor sub-optimal. To make the projection more accurate with respect to Eqn. \ref{eqn:ideal}, Dehaene \textit{et al}. propose to apply adversarial samples generation idea, that is to say, starting from a sample $\mathbf{x}_0$, iterate gradient descent steps over the input $\mathbf{x}$, constructing samples $\mathbf{x}_1, \cdots, \mathbf{x}_N$, to minimize the energy $E(\mathbf{x})$, defined as 
\begin{equation}
	E(\mathbf{x}_t)=\mathcal{L}_r(\mathbf{x}_t) + \lambda\cdot ||\mathbf{x}_t - \mathbf{x}_0||_1
\end{equation}
An iteration is done by calculating $\mathbf{x}_{t+1}$ as
\begin{equation}\label{eqn:proj}
	\mathbf{x}_{t+1} = \mathbf{x}_t - \alpha \cdot \nabla_{\mathbf{x}} E(\mathbf{x}_t),
\end{equation}
where $\alpha$ is the learning rate, and $\lambda$ is a parameter trading off the inclusion of $\mathbf{x}_t$ in the normal manifold, given by $\mathcal{L}_r(\mathbf{x}_t)$, and the proximity between $\mathbf{x}_t$ and the input $\mathbf{x}_0$, assured by the regularization term $||\mathbf{x}_t - \mathbf{x}_0||_1$. This method enables the ``Rec-Error" predictor to significantly outperform all gradient based predictors on simple natural images. However, the effectiveness of this method on more challenging dataset, e.g. brain MRI images, is not clear before our work. To make the notation clear, we call this method ``Proj-Rec-Error".
\par \noindent
\textbf{$\beta$-VAE for better anomaly localization}
As a variant of VAE, $\beta$-VAE \cite{higgins2017beta} is designed for unsupervised discovery of interpretable factorized representations from raw image data. An adjustable hyperparameter $\beta > 1$ is introduced to balance the extent of learning constraints (a limit on the capacity of the latent information and an emphasis on learning statistically independent latent factors) and reconstruction accuracy. Hoffman \textit{et al}. \cite{hoffman2017beta} introduced a reformulation of $\beta$-VAE for $0 < \beta < 1$. They argued that, within this range, training $\beta$-VAE is equivalent to optimizing an approximate log-marginal likelihood bound of VAE under an implicit prior. All in all, different $\beta$ values should induce different balance between latent information (related to $\mathcal{L}_{KL}$) and reconstruction accuracy ($\mathcal{L}_r$), which then change the performance of ``Rec-Error", ``Rec-grad", ``KL-grad" and their combinations. 
\par
Intuitively, a bigger $\beta$, putting less weight on the reconstruction accuracy, may cause insensitive reconstruction change with respect to the input change and then less accurate reconstruction, which is similar to \textit{posterior collapse} \cite{lucas2019don}. The appearance is that a bigger $\beta$ may degenerate the performance of ``Rec-Error". However, the aforementioned argument only makes sense for normal data. On the other side, a bigger $\beta$ encourages more disentangled representation in the latent space and then capture the normal data manifold better, and the corresponding $\beta$-VAE may have superior ability to  ``inpaint'' the abnormal pixels with their corresponding normal pixels successfully, and finally improve the performance of ``Rec-Error". Similar induction can be applied to KL related predictors.
In this work, we will investigate the effects of $\beta$ experimentally.
\par \noindent
\textbf{Predictors ensemble}
We also investigate that whether combining different predictors can improve localization accuracy. Actually, one can think the ``Combi" is one heuristic approach that ensembles the predictor ``KL-grad" and ``Rec-Error". However, it is non-trivial to do this systematically in the fully unsupervised setting. To leverage the full power of all predictors, we have to use a small portion of dataset, which includes both normal and abnormal data, to find the reasonable way to combine all predictors. We propose to utilize a logistic regression model, where the values of different predictors are treated as different features, to ensemble different predictors for possible performance improvement.
\section{Experiments}
In this section, we evaluate the effectiveness of the iterative projection method on more challenging dataset, i.e. the brain MRI images. Later we will test the proposed $\beta$-VAE based anomaly localization method.
\subsection{Dataset}
\textbf{Training dataset:}
To learn the normal brain MRI image distribution, we trained the VAE and $\beta$-VAE on 3T T2 MRI images of Human Connectome Project (HCP) dataset \cite{van2012human}, which are from 1113 healthy young adult (age 22-35) participants. Data augmentation includes random noise adding, random rotation, and color augmentation.
\textbf{Test dataset:}
We evaluate the anomaly localization method on the BraTS2018 dataset \cite{menze2014multimodal,bakas2017advancing,bakas2018identifying}. There are 285 cases in total, and only T2 image of each case was utilized for our experiment. The resolution is 1x1x1 mm isotropic and all image volumes have a size 240x240x155. We do not have access to the BraTS2017 dataset, as the resource link is out of date.
\subsection{Pre-processing and Hyperparameters}
Both training and test dataset were normalized to have zero mean and unit variance, and slice-wise resampled to have a size of 64x64 pixels. 
For training, all models were trained for 500 epochs with an Adam optimizer having an initial learning rate of $10^{-4}$. During inference, for the ``Proj-Rec-Error" method, an Adam optimizer with a learning rate $\alpha=0.03$ in Eqn. \ref{eqn:proj} was utilized.
\subsection{VAE and $\beta$-VAE architecture}
As VAE and $\beta$-VAE differ only on the loss function, for fair comparisons, we set them to have the same architecture as in \cite{zimmerer2019unsupervised}, which consists of a 5 layer fully-convolutinal encoder and decoder with feature-map size of 16-32-64-256. We used strided convolutions (stride 2) as downsampling and transposed convolutions as upsampling operations, each followed by a LeakyReLU non-linearity.

\subsection{Results}
In this section, we aim to answer the following questions: 1) Which predictors are \textit{better}?  2) Is the ``Proj-Rec-Error" effective in medical imaging? 3)  Is a \textit{big} $\beta$ good or bad for anomaly localization? 4) Are these predictors \textit{complementary}? The metric we utilize is the pixle-wise area under a receiver operating characteristic curve (AUROC), which is commonly used for unsupervised binary classification.
\par \noindent
\textbf{Which predictors are better?}
\begin{figure}[tb!]
	\centering
	\includegraphics[width=0.95\textwidth]{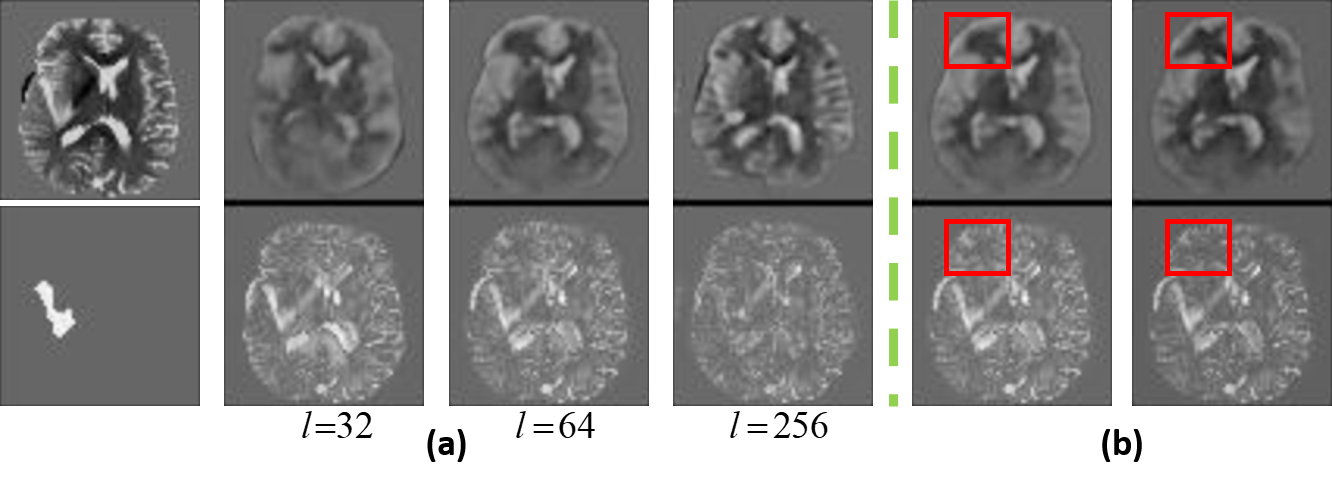}
					\vspace{-0.5cm}
	\caption { Reconstruction of abnormal data. The first column is the input image and the ground truth tumor segmentation. The other columns are reconstructions and their respective reconstruction error images (a) with different latent space dimension sizes; (b) without or with iterative projection.}
	\label{fig:vae_rec}
\end{figure}
\begin{table}[bt!]
	\centering
	\begin{tabular}{|c|c|c|c|c|c|c|}
		\hline
		latent size $l$ & Rec-Error & ELBO-grad & KL-grad & Rec-grad & Combi \\
		\hline
		32 & 0.860 & 0.900 & 0.853 & 0.901 & 0.831 \\
		\hline
		64 & 0.856 & 0.900 & 0.809 & 0.900 & 0.800 \\
		\hline
		256 & 0.838 & 0.871 & 0.853 & 0.871 & 0.803  \\
		\hline
	\end{tabular}
	\caption{AUROC of different predictors using VAE.}
		\label{tab:vae}
		\vspace{-1cm}
\end{table}
As noticed in in \cite{zimmerer2019unsupervised}, the localization performance is highly dependent on two settings of VAE: the image size and the latent space dimension.
The good image size is a trade-off between the modeling difficulty of VAE and the localization accuracy of tumors caused by different resolutions. It is well known that VAEs will encounter difficulty for dataset with large image size. However, if the resolution is too low, the localization will be too coarse. It is non-trivial to select the latent space dimension size $l$, i.e. if it is too big, the learned latent space may not be well defined, e.g. VAE can reconstruct abnormal input successfully, which is demonstrated in Fig.~\ref{fig:vae_rec}(a) denoted by $l=256$; if the size is too small, VAE can not keep all the variational generative factors and then cause over-smoothed reconstruction and hinders accurate abnormal localization, which is demonstrated in Fig.~\ref{fig:vae_rec}(a) denoted by $l=32$. 
\par \noindent
The best settings claimed in \cite{zimmerer2019unsupervised} on the BraTS2017 are: the image size is 64x64; the latent space dimension size is 256. In our experiments on BraTS2018, the scenario of which is similar to that of BraTS2017, we settle down the image size to 64x64 and explore the effect of latent size on the localization performance. The results are listed in the Tab. \ref{tab:vae}. As can be seen, in contradict to the observations in \cite{zimmerer2019unsupervised}, instead of ``KL-grad" and ``Combi", our experiments support that ``Rec-Error", ``ELBO-grad" and ``Rec-grad", are the best predictors, and also the latter two perform comparably for all three latent space dimension sizes.

\par \noindent
\textbf{Effectiveness of ``Proj-Rec-Error" in medical imaging}
%
%
We applied the  ``Proj-Rec-Error", which is highly successful in simple natural images \cite{dehaene2020iterative}, to more challenging brain tumor localization problem. The baseline is the VAE model with latent size 64. The AUROC of ``Proj-Rec-Error" is 0.861, compared to 0.856 of ``Rec-Error" and 0.900 of ``Rec-grad". It can be seen that the projection can boost the performance of ``Rec-error". However, it is still outperformed by the gradient based method ``Rec-grad". But it seems we should blame more on the VAE model, which may not model the normal brain distribution good enough. This can be demonstrated in Fig. \ref{fig:vae_rec}(b), where the iterative projection can correct only part of reconstruction on normal regions, which indicates that the corresponding normal image of the input is not modeled well by the VAE.
\par
\begin{figure}[tb!]
	\centering
	\includegraphics[width=0.7\textwidth]{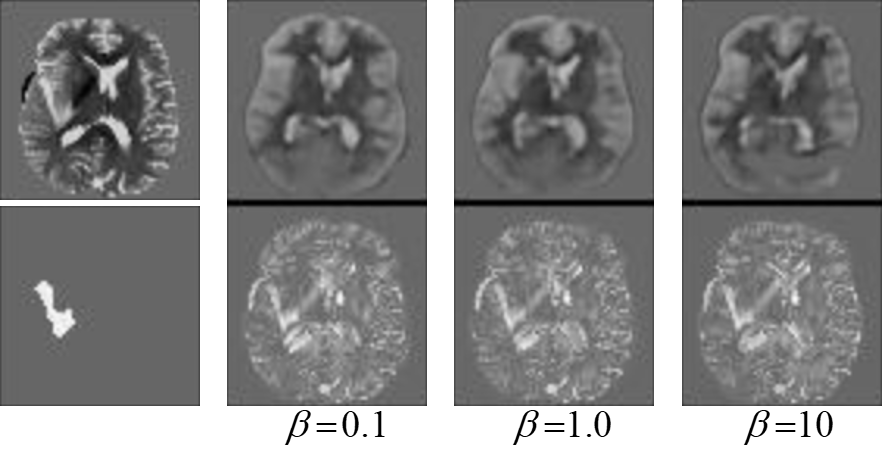}
					\vspace{-0.5cm}
	\caption {$\beta$-VAEs reconstruction of abnormal data.}
	\label{fig:beta_vae_rec}
\end{figure}

 \noindent
\textbf{Does $\beta$-VAE help?}
Based on the indication from the previous part, we explored the modeling capacity of $\beta$-VAEs, which is kind of generalized VAE, and their performance on anomaly localization task. The results using different $\beta$ values are listed in Tab. \ref{tab:beta_vae}. As can be seen, $beta=10$ seems inducing the best performance for the first four predictors, and however interestingly, the performance of ``Combi" degrades consistently as $\beta$ increases. Moreover, ``Proj-Rec-Error" consistently outperforms ``Rec-Error". The reconstruction behavior of $\beta$-VAE is demonstrated in Fig. \ref{fig:beta_vae_rec}. One can notice the reconstruction seems differing from the input as $\beta$ increases. In some sense,  by $\beta$-VAE with bigger $\beta$ for training on normal data distribution, the reconstruction of abnormal data seems more biased to the learned normal data distribution, which is good for abnormal regions but bad for normal regions with respect to Eqn. \ref{eqn:ideal}.
\begin{table}[tb!]
	\centering
	\begin{tabular}{|c|c|c|c|c|c|c|}
		\hline
		$\beta$ & Rec-Error & ELBO-grad & KL-grad & Rec-grad & Combi & Proj-Rec-Error \\
		\hline
		0.1 & 0.857 & 0.899 & 0.805 & 0.899 & 0.810 & 0.861\\
		\hline
		0.5 & 0.856 & 0.903 & 0.809 & 0.903 & 0.808 & 0.865\\
		\hline
		1.0 & 0.856 & 0.900 & 0.809 & 0.900 & 0.800 & 0.861\\
		\hline
		2.0 & 0.856 & 0.900 & 0.803 & 0.900 & 0.778 & 0.862\\
		\hline
		10.0 & 0.859 & 0.905 & 0.810 & 0.905 & 0.749 & 0.864\\
		\hline
	\end{tabular}
	\caption{AUROC of different predictors using $\beta$-VAE.}
	\label{tab:beta_vae}
				\vspace{-0.8cm}
\end{table}
\par \noindent
\textbf{Complementariness of different predictors}
To attempt for further boosting of the localization performance, 10\% test data was utilized to train a logistic regression model using ``Rec-Error", ``KL-grad" and ``Rec-grad" as independent features. We abandon the ``ELBO-grad", since it is just the sum of ``KL-grad" and ``Rec-grad". We use the $\beta$-VAE ($\beta$=10) as the backbone. The resulting weight parameters of the three predictors are in the scale of $10^{-1}, 10^{-1}$ and $10^2$ and the AUROC is $0.903$, which is a little bit worse than that of ``Rec-grad" ($0.905$). This basically indicates that the ``Rec-grad" predictor may include almost all the information within other predictors and they are far from being complementary to each other.
\section{Conclusion}
In this paper, we applied the energy-based projection in more challenging medical imaging scenario and found it is not as useful as on natural images. Moreover, we observe that the robustness of KL gradient  predictor totally depends on the setting of the VAE. We also explored the effect of the weight of KL loss within beta-VAE in anomaly localization. Ensemble of different predictors were also investigated.
%
%
%
\bibliographystyle{splncs04}
%
\bibliography{ref}
\end{document}